\documentclass[10pt,twocolumn]{article}

\usepackage{wacv}
\wacvfinalcopy

\usepackage[utf8]{inputenc}

\usepackage{mathtools}

\usepackage{upgreek}
\usepackage{xspace}

\newcommand{\tm}{\ensuremath{h}\xspace}
\newcommand{\tss}{\ensuremath{H}\xspace}
\newcommand{\tcs}{\ensuremath{H_\upphi}\xspace}
\newcommand{\tpp}{\ensuremath{\phi}\xspace}
\newcommand{\tpf}{\ensuremath{\upphi}\xspace}

\newcommand{\sm}{\ensuremath{g}\xspace}
\newcommand{\sss}{\ensuremath{G}\xspace}
\newcommand{\scs}{\ensuremath{G_\uptheta}\xspace}
\newcommand{\spp}{\ensuremath{\theta}\xspace}
\newcommand{\spf}{\ensuremath{\uptheta}\xspace}

\usepackage{graphicx}
\graphicspath{{imgs/},{results/}}
\usepackage{subcaption}
\usepackage{siunitx}

\newcommand{\x}{\ensuremath{\mathbf{x}} }

\renewcommand{\vec}[1]{\mbox{\boldmath{$#1$}}}

\newcommand{\wmat}[2]{{^{\mathrm{#1}}\vec{W}_{\mathrm{#2}}}}

\renewcommand{\x}{ \ensuremath{\vec{x {} }}}

\renewcommand{\vec}[1]{\mbox{\boldmath{$#1$}}}

\renewcommand{\x}{\vec{x}}
\newcommand{\I}{\mathrm{\vec{I}}}

\newcommand{\Ifr}[1]{\I_{#1}}

\ifwacvfinal
\usepackage[breaklinks=true,bookmarks=false]{hyperref}
\else
\usepackage[pagebackref=true,breaklinks=true,colorlinks,bookmarks=false]{hyperref}
\fi

\ifwacvfinal
\else
\fi

\title{Patient-Specific Domain Adaptation for Fast Optical Flow Based on Student-Teacher Knowledge Transfer}

\author{Sontje Ihler, Max-Heinrich Laves, Tobias Ortmaier\\
Institut für Mechatronische Systeme\\
Leibniz Universität Hannover\\
{\tt\small sontje.ihler@imes.uni-hannover.de}
}

\author{Sontje Ihler, Max-Heinrich Laves, Tobias Ortmaier\\
Institut für Mechatronische Systeme\\
Leibniz Universität Hannover\\
{\tt\small sontje.ihler@imes.uni-hannover.de}
}

\begin{document}
\maketitle

\begin{abstract}
Fast motion feedback is crucial in computer-aided surgery (CAS) on moving tissue. 
Image-assistance in safety-critical vision applications requires a dense tracking of tissue motion. This can be done using optical flow (OF).
Accurate motion predictions at high processing rates lead to higher patient safety.  
Current deep learning OF models show the common speed vs. accuracy trade-off.
To achieve high accuracy at high processing rates, we propose patient-specific fine-tuning of a fast model.
This minimizes the domain gap between training and application data, while reducing the target domain to the capability of the lower complex, fast model.
We propose to obtain training sequences pre-operatively in the operation room.
We handle missing ground truth, by employing teacher-student learning.
Using flow estimations from teacher model FlowNet2 we specialize a fast student model FlowNet2S on the patient-specific domain.
Evaluation is performed on sequences from the Hamlyn dataset.
Our student model shows very good performance after fine-tuning.
Tracking accuracy is comparable to the teacher model at a speed up of factor six.
Fine-tuning can be performed within minutes, making it feasible for the operation room.
Our method allows to use a real-time capable model that was previously not suited for this task. This method is laying the path for improved patient-specific motion estimation in CAS.
\end{abstract}


\section{Introduction}

In robot-assisted minimally invasive surgery (MIS), instruments are inserted through small incisions and observed via video endoscopy.
The remote control of the instruments is unintuitive and requires a trained operator.
Image-guided surgery can help surgeons to operate more safely and accurately.
A challenging open problem in this scenario is visual motion estimation of moving tissue.
Accurate motion predictions with fast feedback rates are crucial to ensure the patient's safety.  
%
%
%
%

Visual motion estimation can be implemented with sparse tracking, for instance based on feature matching, or dense tracking algorithms like optical flow estimation. There is a wide variety of optical flow algorithms, many based on conventional image processing with engineered features (a broad overview is provided in \cite{Sun10}), as well as many recent data-driven approaches \cite{flownet15, flownet2_17, BackToBasics, unflow, sun2018pwc, wulff2015efficient}.

\begin{figure*}[]
	\centering
	\includegraphics[width=0.24\textwidth]{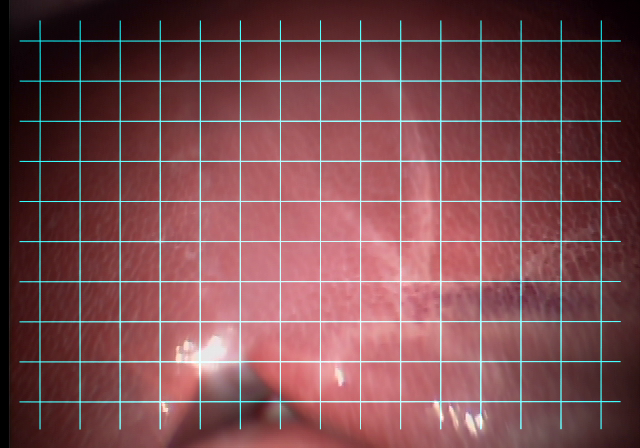}\,%
	\includegraphics[width=0.24\textwidth]{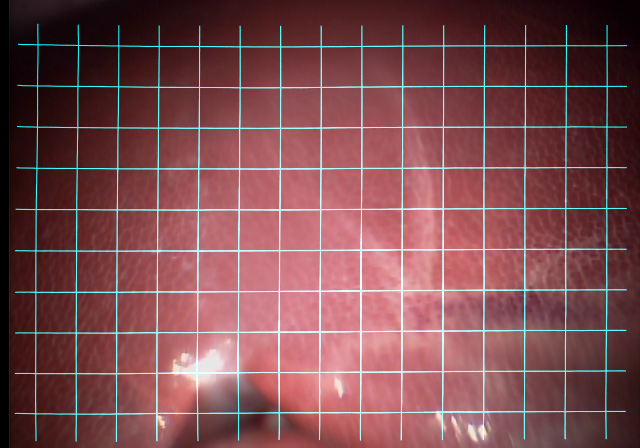}\,%
	\includegraphics[width=0.24\textwidth]{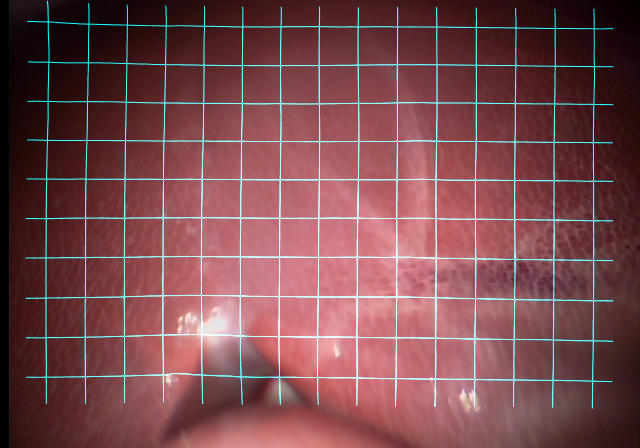}\,%
	\includegraphics[width=0.24\textwidth]{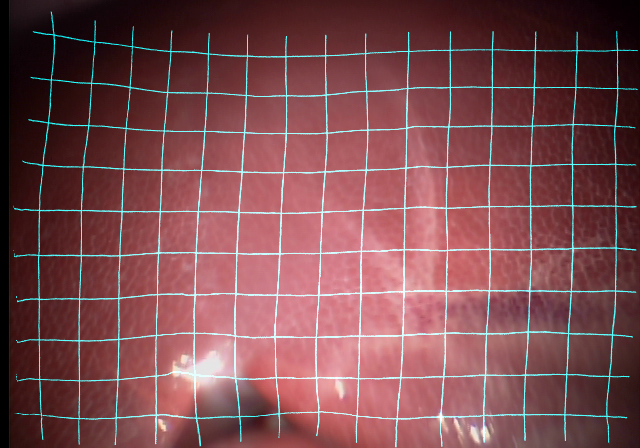}%
	\vspace*{0.06em}
	\includegraphics[width=0.24\textwidth]{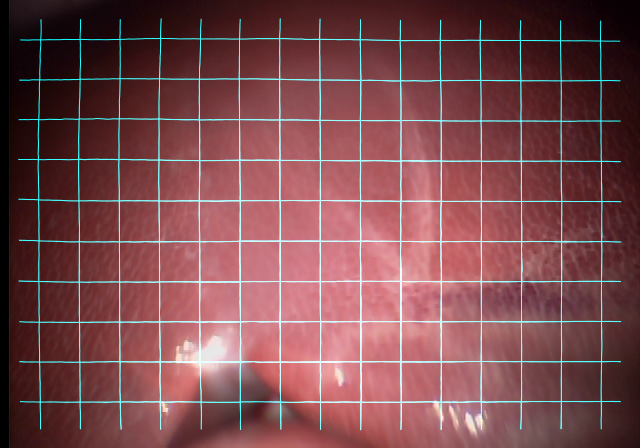}\,%
	\includegraphics[width=0.24\textwidth]{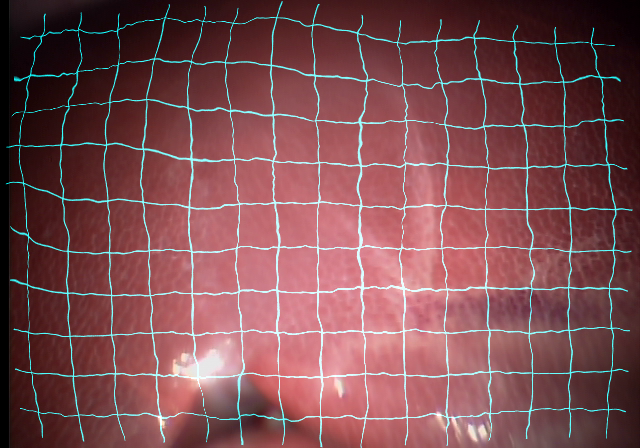}\,%
	\includegraphics[width=0.24\textwidth]{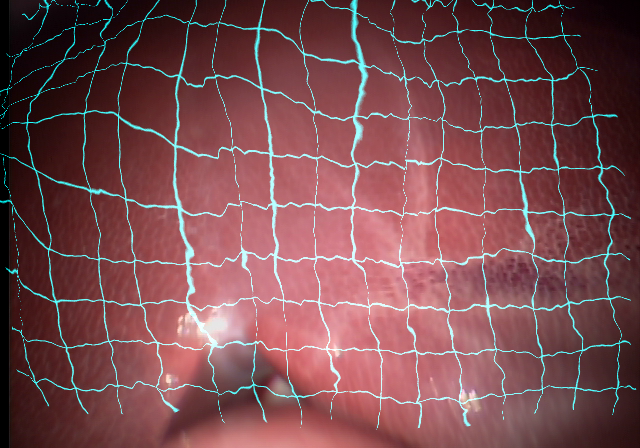}\,%
	\includegraphics[width=0.24\textwidth]{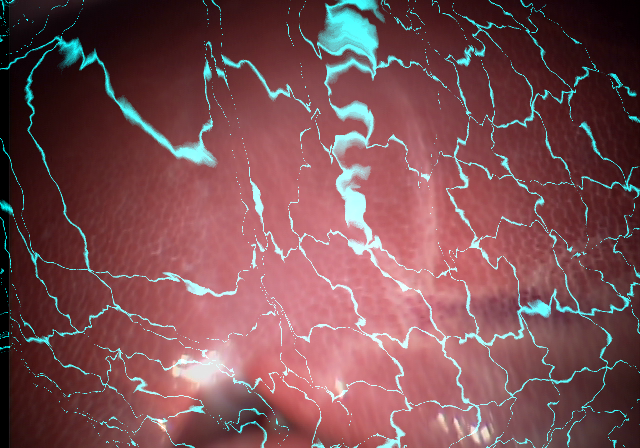}%
	\vspace*{0.06em}
	\includegraphics[width=0.24\textwidth]{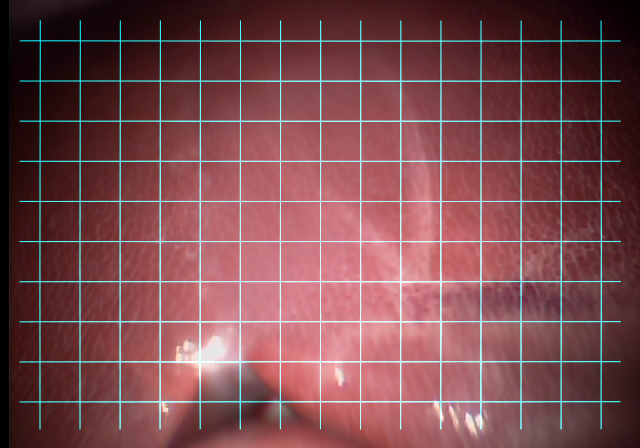}\,%
	\includegraphics[width=0.24\textwidth]{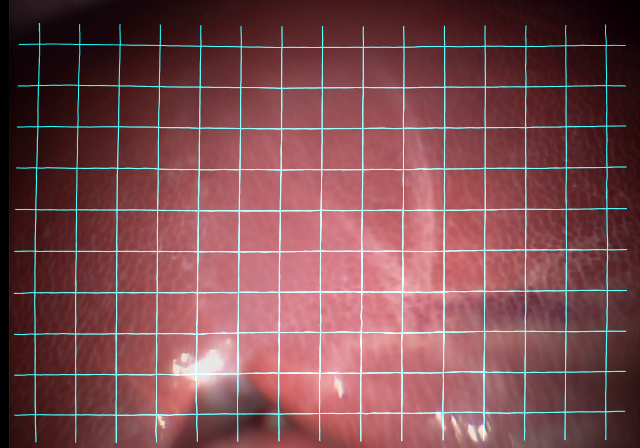}\,%
	\includegraphics[width=0.24\textwidth]{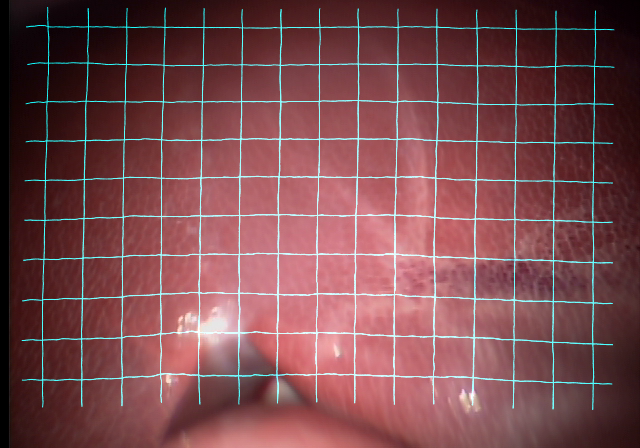}\,%
	\includegraphics[width=0.24\textwidth]{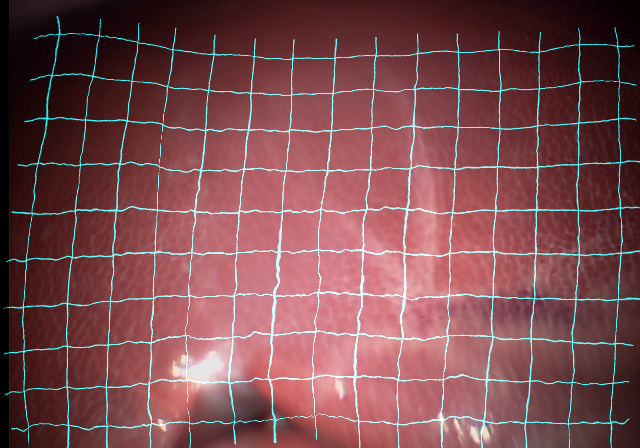}%
	
	\caption{\textbf{Tracking results on a challenging sparse-textured liver} with respiratory motion \cite{HamlynData10} after 0, 12, 38 and 180 frames, respectively from left to right.
		(top row) FlowNet2 @12 fps. (middle row) FlowNet2S @80 fps. (bottom row) our proposed patient-adapted FlowNet2S+F @80 fps.}\label{fig:liver_results}
\end{figure*}

Because tissue deformation can only fully be captured with dense tracking, we focus this work on motion estimation with dense optical flow (OF). We will further focus on deep learning models, as these are outperforming the conventional methods on public OF benchmarks \cite{Sintel12, Geiger2012CVPR, Menze2015CVPR}. Unfortunately, these models show the common speed vs accuracy trade-off. On the one end, they achieve high accuracy in large target domains from high complexity. This leads to slow processing rates. On the other end, faster models lack the capability to generalize well on a large target space and drop in accuracy. For illustration see Figure \ref{fig:liver_results} top and middle row.
Requirements for surgical interventions are accurate motion predictions at simultaneously high processing rates. One might argue that the issue of speed is solved automatically within the next years with the continuous improvement of GPU/TPUs. However, this can only be said for tasks that profit from parallelization, which is not the case for surgical online applications, where each camera frame must be processed right after it is captured.

In this work, we present a combination of patient-specific domain adaption and a teacher-student learning approach to achieve accurate and simultaneously fast OF estimation without ground truth labels.
Patient-specific domain adaptation describes the specialization of a model to a specific patient, in other words, for each patient the OF model is specifically tailored to their individual anatomical motion and appearance. This action maximizes the domain match between the model's training data and the image/motion-data which is afterwards observed during the intervention. This guarantees higher performance of the OF model. To overcome the issue of missing ground truth we deploy a teacher-student-learning strategy for unsupervised domain adaptation. One model functions as a student model. Another model (or several models) function as a teacher model that support the training of the student model. In our case: a complex, accurate, but not real-time-capable OF estimation model serves as a guide for a less complex and fast student model. The complex teacher model operates as a high-performance work horse, which achieves high accuracy in a large task/target space. 
The fast student model lacks the complexity to achieve high accuracy everywhere in large task space, however if the task space is reduced, it can also achieve high accuracy. This is achieved by reducing the task space to the patient-specific task domain. We combine both approaches by capturing video sequences from the situs of planned intervention, use the accurate teacher model to compute a gold truth and fine-tune the fast student model with the gold truth to specialize the student model on the patient-specific image and motion domain.

The contribution of our work is manifold. We propose the concept of patient-specific neural networks to regression tasks without requiring manual annotation and  we introduce patient-specific domain adaptation for motion estimation during surgical interventions. Our method allows to use a real-time capable OF model that was previously not suited for this task. We area able to perform robust optical flow estimation on sparse or deformable tissue at high frame rates.

In the following sections, we first give an overview of related work. We then provide a detailed presentation of proposed unsupervised fine-tuning method. Afterwards we describe the experimental setup to verify our proposed method and show accuracy results on a selection of endoscopic scenarios. The paper concludes with a discussion and outlook.

\section{Related Work}

Optical flow (OF) estimation is a regression problem to determine visible motion between two images. The displacement of corresponding pixels is interpreted as a vector field describing the movement of each image component \cite{HornSchunck81}.
It is widely applied in motion estimation for medical endoscopy: Changes of the camera (endoscope) pose \cite{Mahmoud2017, spyrou2012homography} as well as tissue tracking \cite{Tracking18, Schoob17, TissueTracking2012} can be determined with OF algorithms. Penza et\,al. recently proposed a long-term safety tracking of pre-operatively defined risk areas \cite{Tracking18}. However, the computation time between two frames is to high and does not ensure real-time capability for all speed requiring applications.
In a broader application (especially driven by autonomous driving) deep learning based methods like FlowNet2 \cite{flownet2_17} and PWC-net \cite{sun2018pwc} are outperforming the conventional approaches. Both are designed to be trained in a supervised manner. The lack of ground truth is a general problem in end-to-end learning. FlowNet2 \cite{flownet2_17} profits from training on large synthetically created  (non-medical) image sequences FlyingChairs \cite{flownet15} and FlyingThings3D \cite{flownet2_17}. The image content as well as inter-frame motion in these rendered training sets is drastically different to the properties found in endoscopic surgery. FlowNet2 is a high-complexity network architecture of several stacked subcomponents (FlowNet2S, FlowNet2C, and more), which each for themselves can also function as OF estimation models. As they are smaller, they have much lower inference time but with less accurate performance.  FlowNet2 yields very good motion estimation on endoscopic images, illustrating good general purpose capabilities due to its high complexity. Its small --- and ergo faster --- counterpart FlowNet2S however fails as illustrated in Fig.~\ref{fig:liver_results}.

Tissue tracking is a challenging task as ''endoscopic tissue images do not have strong edge features, are poorly lit and have been limited in providing stable, long-term tracking or real-time processing'' \cite{TissueTracking2012}. The issue of sparse textures was recently tackled  for ocular endoscopy, successfully using a fine-tuned FlowNet2S. Fine-tuning of a pretrained FlowNet2S to retinal images with virtual movement yielded in good motion estimation for mosaicking small retinal scans to a retinal panorama \cite{guerre2018optical}. Retinal images are very challenging due to their low textured image content. The authors state that this was the first time they obtained an estimation accuracy sufficient for this task. To obtain labels for supervised training they implemented virtual motion using an affine motion model (translation and rotation). Affine inter-frame motion can be interpreted as an approximation of camera movement. The high speed capability of FlowNet2S is highly suitable for our application. In pre-experiments,  a fine-tuned FlowNet2S model with virtual affine movement was not able to track tissue deformations, which are more complex than affine motion. An unsupervised fine-tuning approach based on UnFLow \cite{unflow}, which does not have the restriction of a simplified motion model, did not converge on sparse textured image pairs in pre-experiments.

In 2018, Armin et\,al. proposed an unsupervised deep learning method to learn inter-frame correspondences for endoscopic images \cite{armin2018unsupervised}. However, their method is outperformed by the state-of-art model FlowNet2, as well as FlowNet2S (see Appendix~\ref{sec:endoreg}). Lit et\,al. recently proposed DDF-FLow, a teacher-student approach to learn occlusion maps for OF in an unsupervised manner \cite{liu2019ddflow}. Unlike us, they use two models of identical complexity. They do not put their focus on computation speed.



\section{Methods}

We propose a simplified teacher-student learning strategy to create an annotated training set for fine-tuning a real-time capable OF model to specialize on a surgical scene. 

To \textbf{gain speed}, we were inspired by the concept of knowledge distillation (KD) \cite{hinton2015distilling} using a teacher-student learning approach. This strategy was pioneered in model compression \cite{bucilua2006model}, where a small model is trained to imitate a pretrained, complex model (ensemble). It was introduced to train small models for mobile applications. Small models also achieve higher processing rates than their complex counterparts. To achieve high accuracy, we propose patient-specific fine-tuning of the fast model. This has two advantages.
First, we \textbf{minimize the domain gap} between training and application data, enabling higher accuracy during a surgical intervention. Second, patient-specific fine-tuning reduces the target space to the capability of the lower complexity, fast model.
For \textbf{patient-specific fine-tuning} we propose to obtain training sequences intra-operatively in the operation room once the camera is placed in the situs.  
We overcome the issue of missing ground truth annotations, by employing the unsupervised teacher-student learning approach.


\subsection{Gaining Speed}
\begin{figure}[]
	\centering
	\includegraphics[width=0.8\linewidth]{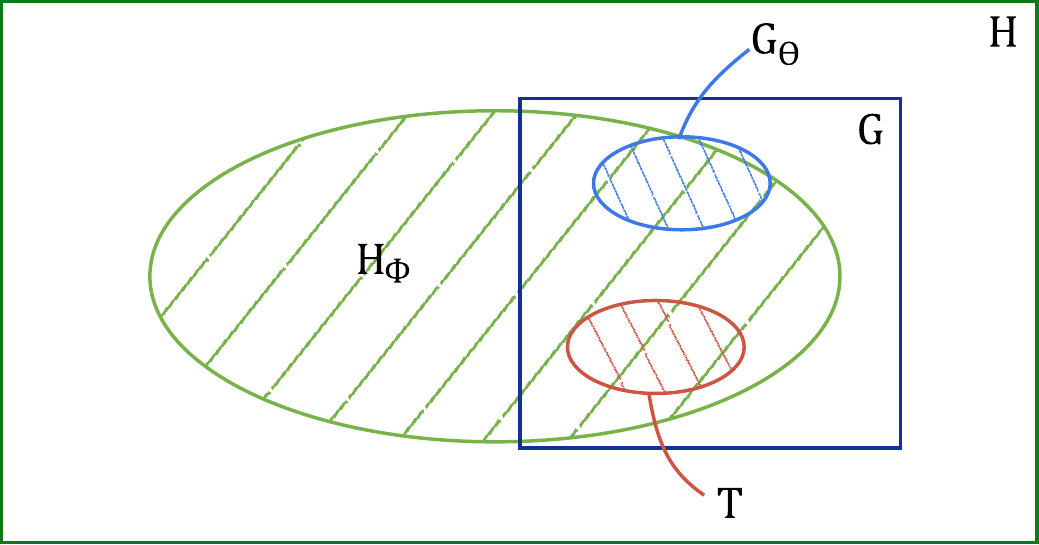}
	\caption{\textbf{Domain adaptation with simplified teacher-student approach.} Goal is to shift the student's convergence space \scs to the target space $T$. In an optimal scenario the target space $T$ fully lies in the  teacher's convergence space \tcs and the student's solution space \sss.}\label{fig:method2}
\end{figure}

To explain our approach we recite a simplified concept of KD model compression, illustrated in Figure~\ref{fig:method2}.

We assume there exists a high-complex teacher model \tm, parametrized by \tpp, with a large solution space \tss and convergence space $\tcs \subset \tss$. The solution space encapsulates all possible mappings that model \tm is capable of, representing the full potential of the model. The convergence space, on the other hand, describes a fixed variant of the model with a fixed parameter configuration \tpf. For neural networks, the parameters are equivalent to the model's weights. 
We now assume a second less-complex, but faster student model \sm. Due its smaller complexity it has a smaller solution space \sss and an even smaller convergence space $\scs \subset \sss$ parametrized with fixed configuration \spf.

The target space $T$ is the domain of the target application (in our case the patient-specific, surgical image/motion domain). The mismatch of the target space with the convergence space of a model can be interpreted as the domain gap. The goal is to shift the student's convergence space to overlap with the target space $T$. If the target space lies within the intersection of the teacher's convergence space and the student's solution space 
\begin{equation}\label{eq:requirements}
	T \subset (\tcs \cap \sss),
\end{equation}
the student's parameter configuration can be altered to replicate the teachers knowledge (convergence space) within the target space.
If the students capacity is great enough to mimick the behavior of the teacher within the target space, is becomes a specialist for the target domain as it achieves the accuracy of the complex model with less computation time.

\subsection{Closing the Gap}

To achieve the shift of the student's convergence space \scs towards the target space $T$,  the objective is to maximize the overlap of \scs and $T$ by finding the optimal solution \spf so that
\begin{equation}
\scs \cap T \xrightarrow{\spp} \textbf{max}.
\end{equation}
This is achieved by fine-tuning the model \sm to the target domain\footnote{Fine-tuning reduces the risk of over-fitting compared to training from scratch. A further advantage is shorter training times (s. Subsection~\ref{sec:patient_specific}))}. It must be noted again, that the target space must be small enough to match the capacity of the student model to achieve high accuracy.  
For fine-tuning, we must sample training samples $(x,y) \in T$ from the target space, where in our case, $x$ are image pairs and $y$ is the corresponding flow field. Unfortunately, in our case it is only possible to sample $x$. Neither corresponding $y$ nor the true mapping $f:x \rightarrow y$ is known. We solve this by employing the mentioned teacher knowledge transfer to realize unsupervised domain adaptation. 

We assume that the requirement in Eq.\ref{eq:requirements} are met and teacher model $\tm:x\rightarrow \tilde{y}$ provides a good approximation for $f$. We can then use \tm to create an annotated dataset $(x, \tilde{y}) \approx (x,y) \in T$ to shift the convergence \scs to overlap with $T$. 
We do that by sampling image pairs  $x$ from the target domain and compute the corresponding flow field $\tilde{y}$ where $\tm(x) = \tilde{y} \approx y$ (gold truth). We can then optimize our student model with the objective $\arg\min_\spp L(\sm_\spp(x), \tilde{y})$.

\subsection{Patient-Specific Target Domain}\label{sec:patient_specific}
Above presented approach requires a small target space to create a fast but accurate student specialist. We already addressed the issue of over-fitting by employing fine-tuning. We also propose to obtain our training samples directly prior to the surgical intervention in the operation room. This reduces the target space efficiently and  simultaneously minimizes the domain gap. To achieve the highest accuracy possible, training samples should at best be identical to application samples. The idea is to capture image data during a preparation stage of an intervention, right after the placement of the camera in the situs. For an optimal fine-tuning outcome of the student model, all expected motion during application should be induced during the caption of training data. 
To make this approach feasible to be performed in the operation room, training times must be very short.

\begin{table*}[h]
	\centering
	\caption{\textbf{Details for datasets and corresponding fine-tuning.} All dataset have a frame rate of 30 fps. Average training time was less than 15 minutes on an Nvidia GeForce GTX 1080 Ti with batch size 8.}
	\label{tab:dataset}
	\resizebox{0.8\textwidth}{!}{%
		\begin{tabular}{ll|ccc|cc}
			\hline
			Dataset & Image Size & \begin{tabular}[c]{@{}c@{}}\# Training \\ Samples\end{tabular} & \begin{tabular}[c]{@{}c@{}}\# Validation\\ Samples\end{tabular} & \begin{tabular}[c]{@{}c@{}}\# Test\\ Samples\end{tabular} & \begin{tabular}[c]{@{}c@{}}Epoch of \\ Convergence\end{tabular} & \begin{tabular}[c]{@{}c@{}}Training\\ Time {[}min{]}\end{tabular} \\ \hline
			rotation & 640x448 & 329 & 110 & 161 & 91 & 12,1 \\
			scale change & 640x448 & 600 & 240 & 397 & 81 & 19,1 \\
			sparse texture & 640x448 & 399 & 100 & 307 & 91 & 14,3 \\
			deformation & 320x512 & 600 & 200 & 100 & 96 & 13,6 \\ \hline
		\end{tabular}%
	}
\end{table*}

\section{Experiments}
The aim of our experiments is to verify that our teacher model is indeed capable of accurate motion estimation in the general endoscopic image domain. We further want to verify that our specialized model has the capacity to achieve high accuracy on the patient-specific target domain. Finally, we compare accuracy of teacher, student and specialist model, as well as the feasibility of our approach as in intra-operative procedure.

\subsection{Dataset}\label{sec:dataset}
To simulate intra-operative sampling from the patient-specific target domain, we use endoscopic video sequences that we split into disjoint training and test sets. The training set represents the sampling phase (preparation stage), while the test set represent the application phase during intervention. 
We chose four endoscopic sequences from the Hamlyn datasets that pose varying challenges regarding image content and inter-frame motion:
\begin{itemize}
	\itemsep0em 
	\item scale and rotation: both in vivo porcine abdomen  \cite{HamlynData10} 
	\item sparse texture (and respiratory motion): in vivo porcine liver \cite{HamlynData10}
	\item strong deformation from tool interaction (and low texture): In vivo lung lobectomy procedure  \cite{HamlynData12}
\end{itemize}
 All sequences show specular lights. The sparse-texture set shows very challenging lighting conditions. The rotation and scale change sequence show artifacts caused by interlacing.
Each dataset represents an individual patient. The splits of the subsets were chosen manually, so that training and test data both cover dataset-specific motion. Exact sectioning of the sequence is shown in the Table~\ref{tab:dataset}. For all datasets the left camera was used.

\subsection{Implementation}
For all our experiments we used the accurate, high-complexity FlowNet2 framework \cite{flownet2_17} as our teacher model \tm  and its fast FlowNet2S component as our low-complexity student model \sm. Due to the very different architectures of the two networks, we focus to mimick only the predictions of the teacher model, instead of learning several behaviors of subcomponents.
Both models have full weight channels and are pretrained on datasets FlyingChairs \cite{flownet15} and FlyingThings3D \cite{MIFDB16}).  We utilized the publicly available implementation in pytorch \cite{FlowNet2-pytorch}. We cropped all datasets to multiples of 64 to fit the models' architecture.


\subsection{Results}
All results were obtained using the test sets described in \ref{sec:dataset}. The test sets were not used during training in any way. The experimental results cover training time, inference time, as well as accuracy of teacher model FlowNet2, student model FlowNet2S and proposed specialist model FlowNet2S+F.

\begin{figure}
	\centering
	\includegraphics[width=0.95\linewidth]{{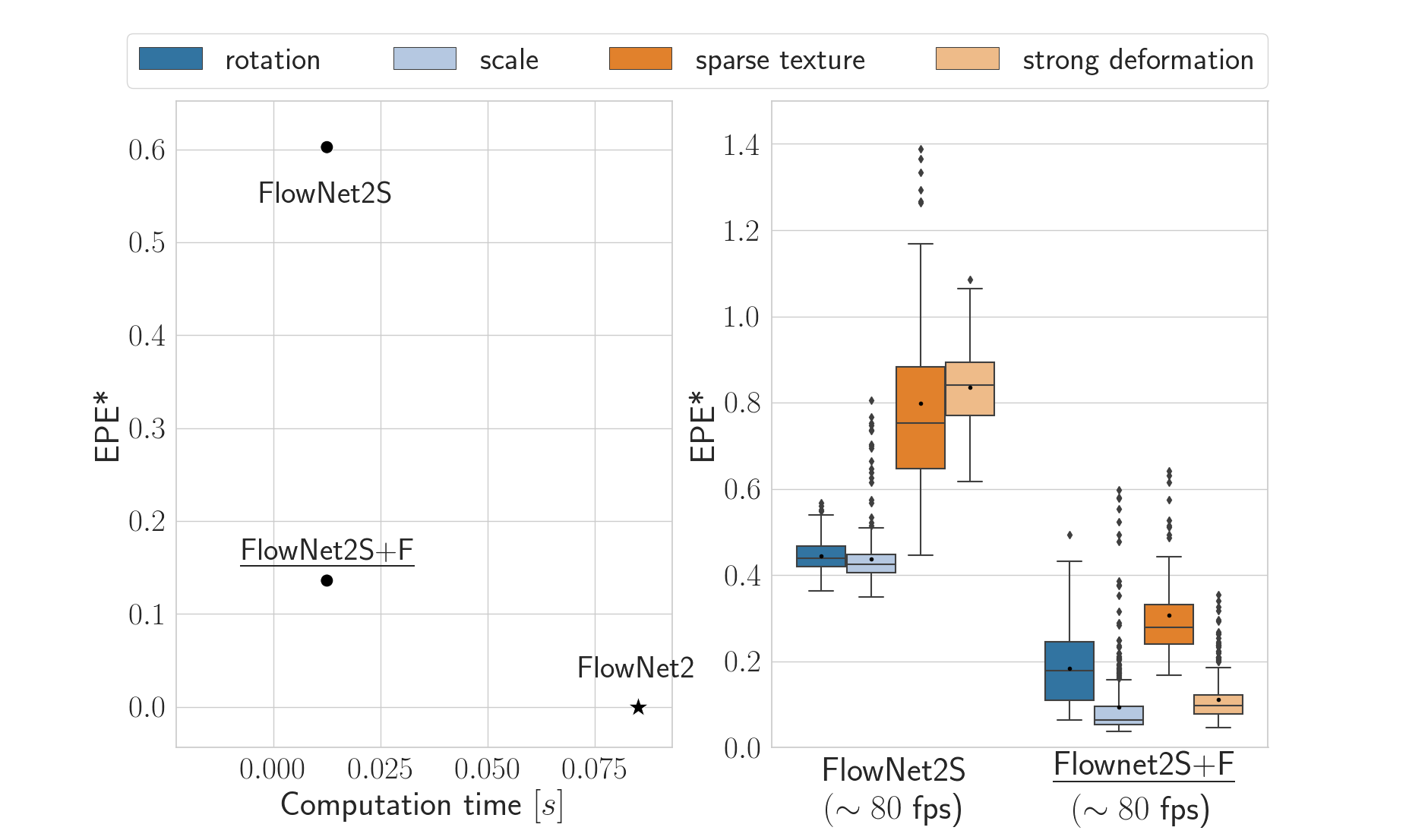}}
	\caption{\textbf{Endpoint error relative to gold truth (EPE*)} --- provided by teacher model FlowNet2 --- for FlowNet2S, as well proposed FlowNet2S+F. The EPE* is provided averaged over all test sets (left) as well as for each set individually (right). The boxplot illustrates the median, lower and upper quartile of all estimation errors, whiskers represent a multiple of 1.5 of the inner quartile range}\label{fig:epe}
\end{figure}

\begin{figure*}
	\centering
	\begin{subfigure}{0.24\textwidth}
		\includegraphics[width=\textwidth]{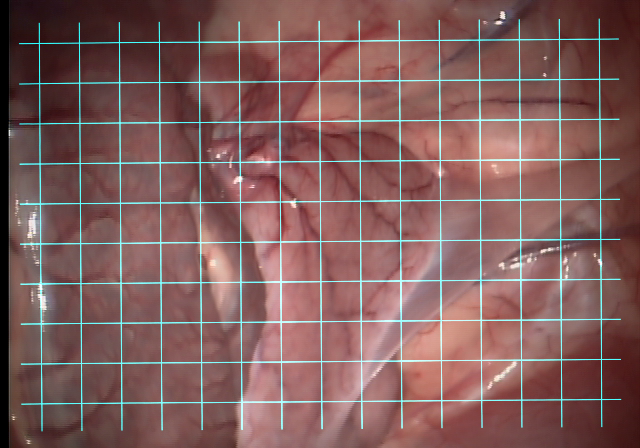}\vspace{0,4em}
		\includegraphics[width=\textwidth]{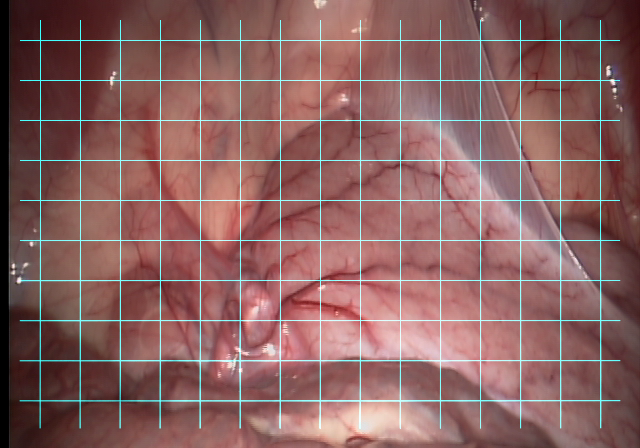}
		\caption*{Prior Tracking}
	\end{subfigure}
		\begin{subfigure}{0.24\textwidth}
		\includegraphics[width=\textwidth]{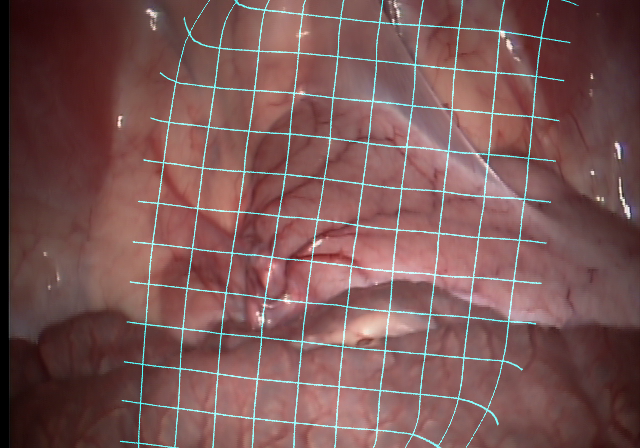}\vspace{0,4em}
		\includegraphics[width=\textwidth]{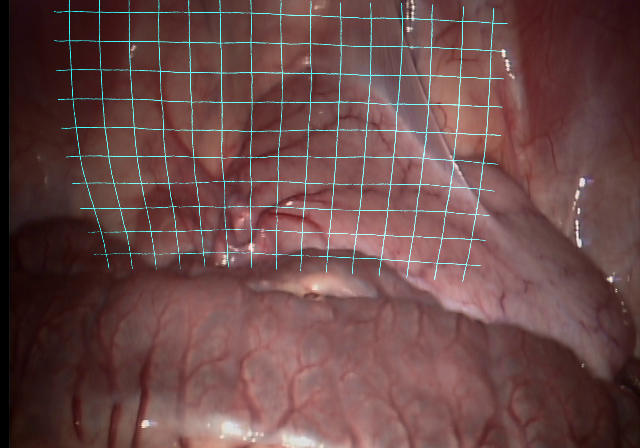}
		\caption*{FlowNet2}
	\end{subfigure}
			\begin{subfigure}{0.24\textwidth}
		\includegraphics[width=\textwidth]{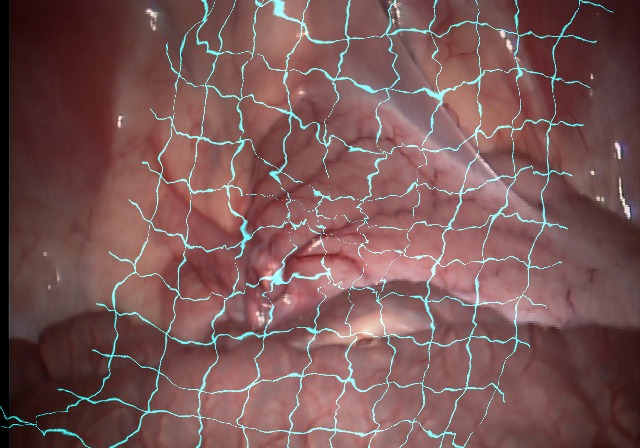}\vspace{0,4em}
		\includegraphics[width=\textwidth]{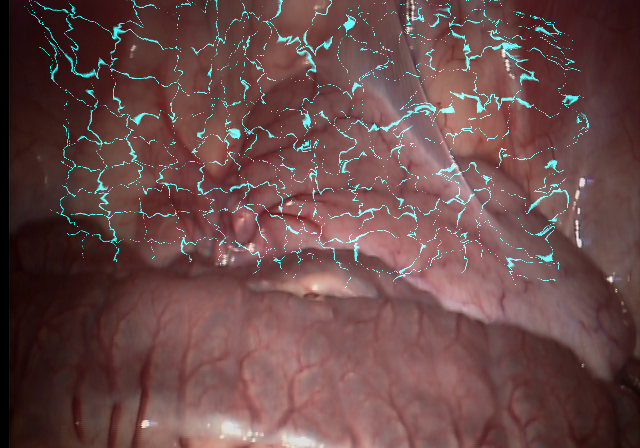}
		\caption*{FlowNet2S}
	\end{subfigure}
			\begin{subfigure}{0.24\textwidth}
		\includegraphics[width=\textwidth]{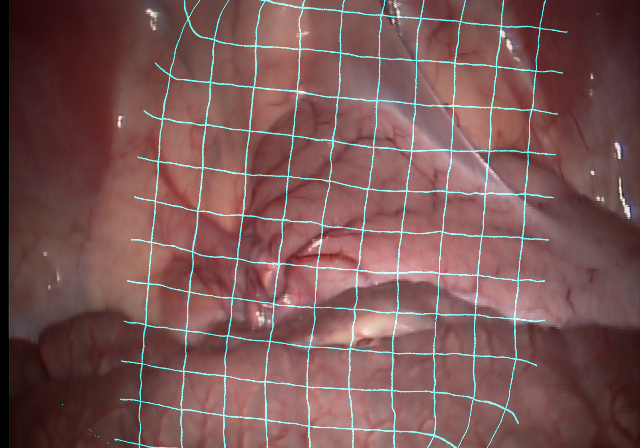}\vspace{0,4em}
		\includegraphics[width=\textwidth]{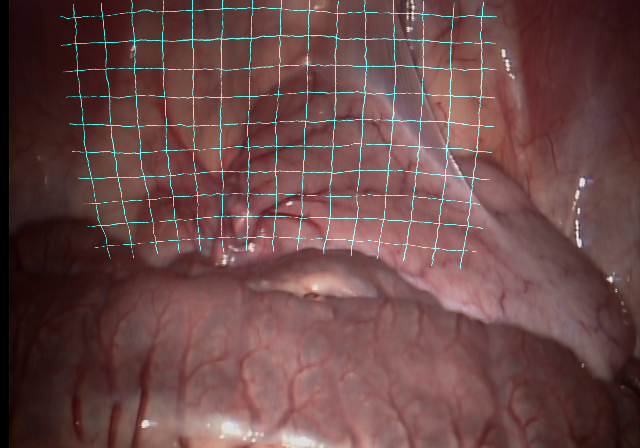}
		\caption*{\underline{FlowNet2SF}}
	\end{subfigure}
\caption{\textbf{Tracking results on rotation (top) and scale (bottom) sequences} from in vivo porcine abdomen \cite{HamlynData10} after 110 (rotation) and 158 frames (scale) consecutive frames. Main challenges is the change of camera pose. There are also specular lights, illumination changes, small respiratory motion, as well as interlacing artifacts.}\label{fig:abdomen_results}
\end{figure*}

\paragraph{Gold truth from teacher model} We compute the gold truth $\tilde{y}=h(x)$ for all data samples $x$ using teacher model FlowNet2 \tm.
\paragraph{Fine-tuning student model} We use student model FlowNet2S pretrained to avoid over-fitting. We followed the (multi-scale) supervised learning scheme proposed by FlowNet2 with loss function $L(\sm(x), \tilde{y}) = ||\sm(x)- \tilde{y})||_1$. Training was performed on training set described in Subsection~\ref{sec:dataset}. We split our training sets into training and validation set (see Table~\ref{tab:dataset}). Fine-tuning was performed until convergence of validation loss, up to a maximum of 100 epochs. Validation loss was computed every five epochs. During training, we augmented the training samples using random crops. Training parameters can be found in Appendix~\ref{sec:training_params}.

\paragraph{Training time} Training times at point of convergence are provided in Table~\ref{tab:dataset}. Average training was less than 15 minutes on an Nvidia GeForce GTX 1080 Ti with batch size 8. Unlike online inference where only single images can be processed at once, training time can further be reduced from parallelization and be accelerated with higher batch sizes (higher memory GPU). This makes training possible within minutes. We therefore consider the proposal of intra-operative sampling of training data feasible.

	

\begin{figure*}
		\begin{subfigure}{0.24\textwidth}
			\includegraphics[width=\textwidth]{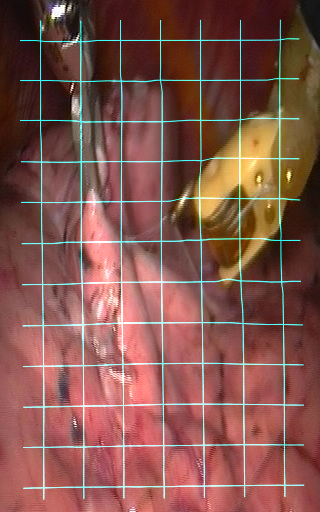}%
			\caption*{Prior Tracking}
		\end{subfigure}
				\begin{subfigure}{0.24\textwidth}
			\includegraphics[width=\textwidth]{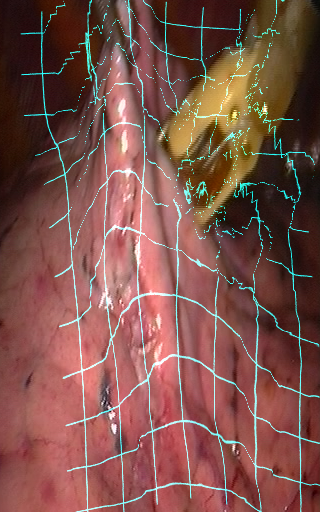}%
			\caption*{FlowNet2}
		\end{subfigure}
				\begin{subfigure}{0.24\textwidth}
			\includegraphics[width=\textwidth]{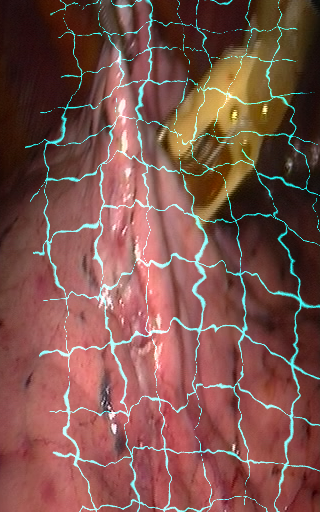}%
			\caption*{FlowNet2S}
		\end{subfigure}
				\begin{subfigure}{0.24\textwidth}
			\includegraphics[width=\textwidth]{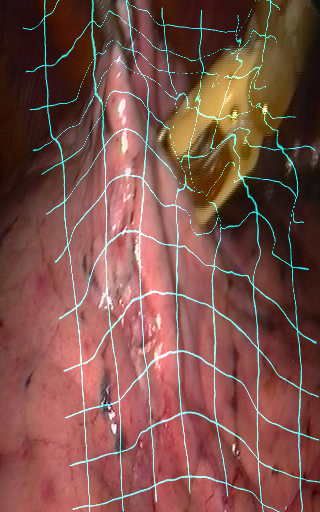}%
			\caption{\underline{FlowNet2S+F}}
		\end{subfigure}
			\caption{\textbf{Tracking results on strong tissue deformation from tool interaction} on low-textured tissue after 60 consecutive frames \cite{HamlynData12}. }\label{fig:deformationresults}
\end{figure*}

\paragraph{Accuray and speed} We provide the relative endpoint error ($\text{EPE*} = ||(g(x)- \tilde{y})||_2$) as well inference time in Figure~\ref{fig:epe} on all test sets for models FlowNet2, FlowNet2S and our proposed model FlowNet2S+F. Inference time of FlowNet2 was \SI{0.083}{\second} (processing rate: approx. \SI{12}{fps}) on our dataset, while FlowNet2S is significantly faster at  \SI{0.013}{\second}, (processing rate approx. \SI{80}{fps}). Our fine-tuned model FlowNet2S+F has identical inference time as FlowNet2S. FlowNet2S+F is therefore 6 times as fast as its complex counterpart FlowNet2.

We show EPE* results averaged over the entire sequence, as well as results for each test sequence separately to see how FlowNet2S and our proposed model FlowNet2S+F handles rotation, scale, sparse texture and deformation. To show outliers and maximum errors, we chose boxplots rather than the common average. The number of samples in each boxplot is provided in Table~\ref{tab:dataset}. As we employed FlowNet2 as the benchmark to compute EPE*, the error of FlowNet2 naturally is zero.

The fine-tuned model produces optical flow estimations much closer to the ones estimated from FlowNet2. Initial FlowNet2S showed an initial average EPE* of 0.6 which was reduced to 0.12. Generally, sparse texture and deformation seem more challenging for FlowNet2S than rotation and scale change. After fine-tuning, proposed FlowNet2S+F achieves comparable accuracy to FlowNet2 on the deformation test set. The sparse texture seems to pose the biggest challenge also for FlowNet2S+F\footnote{We propose an explanation in the Appendix~\ref{sec:illu_issues}}, nevertheless, the error is still reduced to less than half. Overall, both models achieve low, relative errors (below \SI{1}{px}) on all test sets. However, this does not imply, their the estimations are accurate enough for tracking. As already shown in Figure~\ref{fig:liver_results} FlowNet2S fails on this task.


\paragraph{Tracking} Due to the lack of annotated ground truth, we cannot evaluate the estimation accuracy on the test sequences directly. To overcome this issue, we embed the flow estimation into a tracking algorithm over consecutive frames. Small errors between two consecutive frames are not visible. During tracking small errors add up for consecutive frame, resulting in drift and making small errors visible over time.

We illustrate our tracking results in Figures~\ref{fig:liver_results}, \ref{fig:abdomen_results}~and~\ref{fig:deformationresults}.
 We compare estimation accuracy of our fine-tuned model FlowNet2S+F to existing models FlowNet2 and FlowNet2S.
The long-term tracking based on FlowNet2 all yield robust tracking results on all our test sets, confirming its capability as a suitable teacher model. There is a small occurrence of drift, visible in Figure~\ref{fig:abdomen_results} bottom row after 158 frames. Before fine-tuning, the student model FlowNet2S fails on all test set. After fine-tuning however, our customized FlowNet2S+F model mimicks the predictions of FlowNet2 very well, with few exceptions at the margins. We deduce that FlowNet2S is in fact capable to learn highly-accurate flow estimation for small, patient-specific target domains.
Overall, our specialist  model achieves comparable accuracy on all test sets compared to the complex FlowNet2 framework, which is more than factor 6 slower in inference time.

\section{Conclusion}

In this work we introduced a novel method to create a patient-adaptive OF algorithm on-the-fly during pre-operation setup based on the concept of patient-specific vision algorithms.
Our method allows us to use a real-time capable model that was previously not suitable for this task. 

For evaluation of our method, we used  in vivo endoscopic video sequences from the Hamlyn Dataset \cite{HamlynData10, HamlynData12}. We created gold truth annotations for our training data using FlowNet2 as a high-performance teacher model. The gold truth was used to fine-tune FlowNet2S, a simple, therefore real-time capable student model \cite{flownet2_17}. We benchmarked our approach by comparing FlowNet2 (accurate but slow) and FlowNet2S (fast but less accurate) embedded in a tissue tracking algorithm, as well as providing the relative EPE for the fast models. Overall, our specialist model FlowNet2S+F achieves comparable estimation accuracy to the complex FlowNet2 framework on our test set at a significant speed up (more than 6 times faster).

Our fine-tuning method significantly reduced the issue of drift for model FlowNet2S on our test set, making it feasible for robust long-term tissue tracking at high processing speeds. 
Inference time on our test images was only \SI{13}{\milli\second}. This not only enables the processing of higher camera frame rates up to \SI{80}{fps} increasing the update rate of estimated motion rates but also generally reduces computation delay for slower frame rates. Overall, high processing rates improve the safety of vision-assisted applications.

With fine-tuning taking less than 15 minutes on average, the proposed method can be used as a preprocessing step before operation. The method is laying the path for improved patient-specific motion estimation and tracking for computer aided surgery.
Overall, we demonstrate the feasibility of our training scheme. We achieved comparable accuracy and robustness to drift to state of the art model architecture for accurate OF at a significant speed up, making it feasible for real-time online application. Faster processing rates result in faster tracking feedback and can significantly increase the safety of a surgical intervention. 

In our experiments, we only tested our method on FlowNet2S, however, any adequate end-to-end convnet for OF is feasible. Future experiments could also be done on FlowNets, a sparse version (sparse weights) of FlowNet2S. At the expense of possibly reduced accuracy, it is possible to achieve even faster flow computation. It would further be interesting to teach the student model using a teacher ensemble, fully exploiting the potential of distilled knowledge. This has the further advantage of introducing predictive uncertainty to our motion estimation. The uncertainty of a prediction has significant influence on the safety of a vision application. For real-life application the method would also have to be extended by occlusion handling.

\section*{Acknowledgements}
	This research has received funding from the European Union as being part of the EFRE OPhonLas project.\\
	

{\small
\bibliographystyle{apalike}
\bibliography{main}
}

\appendix
\section{Training Parameters}\label{sec:training_params}
added soon

\section{Tracking algorithm}
Image Stabilization (or motion compensation) based on estimated flow fields $\wmat{i}{i+1}$ is achieved by warping each frame $\Ifr{n}$ to reconstruct the first frame $\hat{\I}_1$.
\begin{align}
\hat{\I}_1(\x) &= \Ifr{n}(\x + \wmat{1}{n}(\x))\\ 
&= \Ifr{n}(\x + \underbrace{\wmat{1}{2}(\wmat{2}{3}(... (\wmat{n-1}{n}}_{\text{flow concatenation}}(\x))...)
\end{align}
Tracking is the inverse operation of image stabilization.

\begin{figure}
	\centering
	\includegraphics[width=0.8\linewidth]{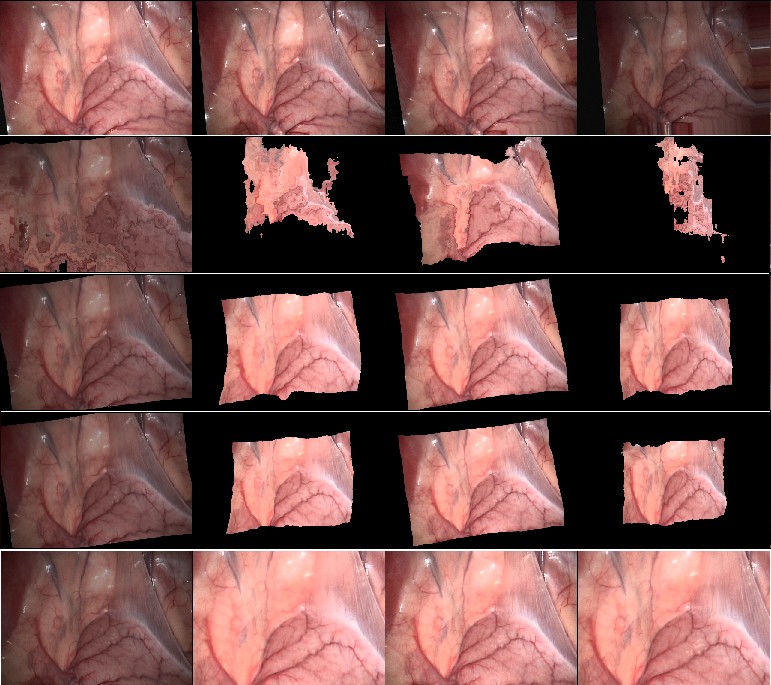}
	\caption{Progress of Training FlowNet2S with illumination augmentation. Showing flow-based image reconstructions before epoch 1, 2 and 3. Top and bottom are initial image pairs. Second, third and fourth row show training progress before epoch 1, 2 and 3 respectively.  It shows that FlowNet2S cannot handle strong illumination changes well without fine-tuning.}\label{fig:illu_challenge}
\end{figure}

\section{FlowNet2S and its struggle with illumination changes}\label{sec:illu_issues}
It is not only sparse texture that is challenging in the liver dataset of this work, but also very difficult lighting condition around the border. The shading seems like vignetting, but is actually due to shape of liver in combination with frontal illumination. As a result transition from light to dark areas should not be used as a reference for flow estimation. A well-working OF model would require invariance to illumination. As can be seen in Figure~\ref{fig:illu_challenge}, FlowNet2S does not handle strong illumination changes well. This should definitely be subject to future work.

\section{EndoRegNet}\label{sec:endoreg}
Armin et.\,al propose a method called EndoRegNet for estimating flow between two endoscopic frame using an unsupervised learning scheme \cite{armin2018unsupervised}. They provide the structural similarity index (SSIM) for their method on the rotation and scale dataset also used in this work. For comparison, we computed the same measure derived from FlowNet2 and FlowNet2S on our test sets, which both perform significantly better than EndoRegNet, see Table~\ref{tab:endoreg}

\begin{table}[h]
	\centering
	\caption{Structural Similarity Index on Hamlyn rotation and scale dataset. *Evaluation of FlowNet2 and FlowNet2S was performed on a subset (test set of this work). }
	\label{tab:endoreg}{%
		\begin{tabular}{lrrrrl}
			\hline
			SSIM & \multicolumn{1}{l}{EndoRegNet} & \multicolumn{1}{l}{FlowNet2} & \multicolumn{1}{l}{FlowNet2-S} & \multicolumn{1}{l}{} &  \\ \hline
			mean & $\sim$0.85 & 0.9615* & 0.9551* &  &  \\
			min & $\sim$0.83 & 0.8785* & 0.8650* &  &  \\
			max & $\sim$0.87 & 0.9894* & 0.9845* &  &  \\ \hline
		\end{tabular}%
	}
\end{table}

\end{document}